\documentclass{article} 
\usepackage{iclr2016_conference,times}
\usepackage{url}
\usepackage{latexsym}

\usepackage{times}
\usepackage{amsmath}
\usepackage{amsthm}
\usepackage{amssymb}
\usepackage{graphicx}
\usepackage{xspace}
\usepackage{tabularx}
\usepackage{multicol}
\usepackage{multirow}
\usepackage{url}
\usepackage{wrapfig}
\usepackage{comment}
\usepackage{listings}
\usepackage{color}
\usepackage[utf8]{inputenc}
\usepackage{fancyvrb}
\usepackage{booktabs}
\usepackage{color}
\usepackage[normalem]{ulem}

\newcommand{\bmx}[0]{\begin{bmatrix}}
\newcommand{\emx}[0]{\end{bmatrix}}

\newcommand{\vect}[1]{\mathbf{#1}}
\newcommand{\vects}[1]{\boldsymbol{#1}}
\newcommand{\matr}[1]{\mathbf{#1}}

\newcommand{\vb}[0]{\vect{b}}
\newcommand{\vc}[0]{\vect{c}}

\newcommand{\vh}[0]{\vect{h}}

\newcommand{\vx}[0]{\vect{x}}
\newcommand{\vz}[0]{\vect{z}}
\newcommand{\vw}[0]{\vect{w}}
\newcommand{\vs}[0]{\vect{s}}
\newcommand{\vf}[0]{\vect{f}}
\newcommand{\vi}[0]{\vect{i}}
\newcommand{\vo}[0]{\vect{o}}

\newcommand{\vr}[0]{\vect{r}}
\newcommand{\vp}[0]{\vect{p}}
\newcommand{\mW}[0]{\matr{W}}
\newcommand{\mP}[0]{\matr{P}}

\newcommand{\mE}[0]{\matr{E}}

\newcommand{\mU}[0]{\matr{U}}

\newcommand{\TT}[0]{\vects{\theta}}

\newcommand{\LL}[0]{\mathcal{L}}

\newcommand{\RR}[0]{\mathbb{R}}

\title{Larger-Context Language Modelling \\ with Recurrent Neural Network}

\author{Tian Wang \\
Center for Data Science \\
New York University \\
\texttt{t.wang@nyu.edu} \\
\And
Kyunghyun Cho \\
Courant Institute of Mathematical Sciences \\ 
and Center for Data Science \\
New York University \\
\texttt{kyunghyun.cho@nyu.edu}
}

\begin{document}

\maketitle

\begin{abstract}

In this work, we propose a novel method to incorporate corpus-level discourse
information into language modelling. We call this larger-context language model. 
We introduce a late fusion approach to a recurrent language model based on long 
short-term memory units (LSTM), which helps the LSTM unit keep intra-sentence 
dependencies and inter-sentence dependencies separate from each other.
Through the evaluation on three corpora (IMDB, BBC, and Penn TreeBank), we
demonstrate that the proposed model improves perplexity significantly. In the 
experiments, we evaluate the proposed approach while varying the number of 
context sentences and observe that the proposed late fusion is superior to the 
usual way of incorporating additional inputs to the LSTM. By analyzing the trained 
larger-context language model, we discover that content words, including nouns, 
adjectives and verbs, benefit most from an increasing number of context sentences. 
This analysis suggests that larger-context language model improves the unconditional 
language model by capturing the theme of a document better and more easily.
\end{abstract}

\section{Introduction}

The goal of language modelling is to estimate the probability distribution of
various linguistic units, e.g., words, sentences \citep{rosenfeld2000}. Among
the earliest techniques were count-based $n$-gram language models which intend
to assign the probability distribution of a given word observed after a fixed
number of previous words. Later \citet{bengio2003neural} proposed feed-forward
neural language model, which achieved substantial improvements in perplexity
over count-based language models. \citeauthor{bengio2003neural} showed that
this neural language model could simultaneously learn the conditional
probability of the latest word in a sequence as well as a vector representation
for each word in a predefined vocabulary. 

Recently recurrent neural networks have become one of the most widely used
models in language modelling \citep{mikolov2010recurrent}. Long short-term
memory unit \citep[LSTM,][]{hochreiter1997long} is one of the most common
recurrent activation function. Architecturally speaking, the memory state and
output state are explicitly separated by activation gates such that the
vanishing gradient and exploding gradient problems described in
\citet{bengio1994learning} is avoided. Motivated by such gated model, a number
of variants of RNNs (e.g. \citet[GRU,][]{cho2014learning},
\citet[GF-RNN,][]{junyoung2015gated}) have been designed to easily capture
long-term dependencies. 

When modelling a corpus, these language models 
assume the mutual independence among sentences, and the task is often reduced 
to assigning a probability to a single sentence. In this work, 
we propose a method to incorporate corpus-level discourse dependency into neural 
language model. We call this larger-context language model. It models the influence of context 
by defining a conditional probability in the form of $P(w_{n}|w_{1:n-1},S)$, 
where $w_{1},...,w_{n}$ are words from the same sentence, and $S$ represents 
the context which consists a number of previous sentences of arbitrary length. 

We evaluated our model on three different corpora (IMDB, BBC, and Penn TreeBank). 
Our experiments demonstrate that the proposed larger-context language model improve perplexity
for sentences, significantly reducing per-word perplexity compared to the
language models without context information. Further, through Part-Of-Speech
tag analysis, we discovered that content words, including nouns, adjectives and
verbs, benefit the most from increasing number of context sentences. Such
discovery led us to the conclusion that larger-context language model improves
the unconditional language model by capturing the theme of a document. 

To achieve such improvement, we proposed a late fusion approach, which is a
modification to the LSTM such that it better incorporates the discourse context
from preceding sentences. In the experiments, we evaluated the proposed
approach against early fusion approach with various numbers of context
sentences, and demonstrated the late fusion is superior to the early fusion
approach. 

Our model explores another aspect of context-dependent recurrent language
model. It is novel in that it also provides an insightful way to feed
information into LSTM unit, which could benefit all encoder-decoder based
applications.

\section{Background: Statistical Language Modelling}

Given a document $D=(S_1, S_2, \ldots, S_L)$ which consists of $L$ sentences,
statistical language modelling aims at computing its probability $P(D)$. It is
often assumed that each sentence in the whole document is mutually independent
from each other:
\begin{align}
    \label{eq:corpus_lm}
    P(D) \approx \prod_{l=1}^L P(S_l).
\end{align}
We call this probability (before approximation) a {\em corpus-level
probability}. Under this assumption of mutual independence among sentences, the
task of language modelling is often reduced to assigning a probability to a
single sentence $P(S_l)$. 

A sentence $S_l=(w_1, w_2, \ldots, w_{T_l})$ is a variable-length sequence of
words or tokens. By assuming that a word at any location in a sentence is
largely predictable by preceding words, we can rewrite the sentence probability
into
\begin{align}
    \label{eq:rnn_lm}
    P(S) = \prod_{t=1}^{T_l} p(w_t | w_{<t}),
\end{align}
where $w_{<t}$ is a shorthand notation for all the preceding words. We call this
a {\em sentence-level probability}.

This rewritten probability expression can be either directly modelled by a
recurrent neural network \citep{mikolov2010recurrent} or further approximated as
a product of $n$-gram conditional probabilities such that
\begin{align}
    \label{eq:ngram_lm}
    P(S) \approx \prod_{t=1}^{T_l} p(w_t | w_{t-n}^{t-1}),
\end{align}
where $w_{t-n}^{t-1} = (w_{t-n}, \ldots, w_{t-1})$. The latter is called
{\em $n$-gram language modelling}. See, e.g., \citep{kneser1995improved} for detailed
reviews on the most widely used techniques for $n$-gram language modelling.

The most widely used approach to this statistical language modelling is $n$-gram
language model in Eq.~\eqref{eq:ngram_lm}. This approach builds a large table of
$n$-gram statistics based on a training corpus. Each row of the table contains
as its key the $n$-gram phrase and its number of occurrences in the training
corpus. Based on these statistics, one can estimate the $n$-gram conditional
probability (one of the terms inside the product in Eq.~\eqref{eq:ngram_lm}) by
\begin{align*}
    p(w_t | w_{t-n}^{t-1}) = \frac{c(w_{t-n}, \ldots, w_{t-1}, w_t)}
    {\sum_{w' \in V} c(w_{t-n}, \ldots, w_{t-1}, w')},
\end{align*}
where $c(\cdot)$ is the count in the training corpus, and $V$ is the vocabulary
of all unique words/tokens. As this estimate suffers severely from data
sparsity (i.e., most $n$-grams do not occur at all in the training corpus), many
smoothing/back-off techniques have been proposed over decades. One of the most
widely used smoothing technique is a modified Kneser-Ney
smoothing~\citep{kneser1995improved} 

More recently, \citet{bengio2003neural} proposed to use a feedforward neural
network to model those $n$-gram conditional probabilities to avoid the issue of
data sparsity. This model is often referred to as {\em neural language model}.

This $n$-gram language modelling is however limited due to the $n$-th order
Markov assumption made in Eq.~\eqref{eq:ngram_lm}. Hence,
\citet{mikolov2010recurrent} proposed recently to use a recurrent neural network
to directly model Eq.~\eqref{eq:rnn_lm} without making any Markov assumption. We
will refer to this approach of using a recurrent neural network for language
modeling as {\em recurrent language modelling}.

A recurrent language model is composed of two function--transition and output
functions. The transition function reads one word $w_t$ and updates its hidden
state such that
\begin{align}
    \label{eq:rnn_trans}
    \vh_t = \phi\left( w_t, \vh_{t-1} \right),
\end{align}
where $\vh_0$ is an all-zero vector. $\phi$ is a recurrent activation function,
and two most commonly used ones are long short-term memory units
\citep[LSTM,][]{hochreiter1997long} and gated recurrent units
\citep[GRU,][]{cho2014learning}. For more details on these recurrent activation
units, we refer the reader to \citep{jozefowicz2015empirical,greff2015lstm}.

At each timestep, the output function computes the probability over all possible
{\em next} words in the vocabulary $V$. This is done by
\begin{align}
    \label{eq:rnn_out}
    p(w_{t+1} = w'|w_1^{t}) \propto \exp\left(g_{w'}(\vh_t)\right).
\end{align}
$g$ is commonly implemented as an affine transformation:
\begin{align*}
    g(\vh_t) = \mW_o \vh_t + \vb_o,
\end{align*}
where $\mW_o \in \RR^{|V| \times d}$ and $\vb_o \in \RR^{|V|}$. 

The whole model is trained by maximizing the log-likelihood of a training corpus
often using stochastic gradient descent with backpropagation through time
\citep[see, e.g.,][]{rumelhart1988learning}.

These different approaches to language modelling have been extensively tested
against each other in terms of speech recognition and machine translation in
recent years
\citep{sundermeyer2015feedforward,baltescu2014pragmatic,schwenk2007continuous}.
Often the conclusion is that all three techniques tend to have different
properties and qualities dependent on many different factors, such as the size
of training corpus, available memory and target application. In many cases, it
has been found that it is beneficial to combine all these techniques together
in order to achieve the best language model.

One most important thing to note in this conventional approach to statistical
language modelling is that every sentence in a document is assumed independent
from each other (see Eq.~\eqref{eq:corpus_lm}.) This raises a question on how
strong an assumption this is, how much impact this assumption has on the
final language model quality and how much gain language modelling can get by
making this assumption less strong.

\subsection{Language Modelling with Long Short-Term Memory}
\label{sec:lstm_lm}

Here let us briefly describe a long short-term memory unit which is widely used
as a recurrent activation function $\phi$ (see Eq.~\eqref{eq:rnn_trans}) for
language modelling~\citep[see, e.g.,][]{graves2013generating}.

A layer of long short-term memory (LSTM) unit consists of three gates and a single
memory cell. Three gates--input, output and forget-- are computed by
\begin{align}
    \label{eq:lstm_gates}
    \vi_t =& \sigma\left( \mW_i \vx_t + \mU_i \vh_{t-1} + \vb_i \right) \\
    \vo_t =& \sigma\left( \mW_o \vx_t + \mU_o \vh_{t-1} + \vb_o \right) \\
    \vf_t =& \sigma\left( \mW_f \vx_t + \mU_f \vh_{t-1} + \vb_f \right),
\end{align}
where $\sigma$ is a sigmoid function. $\vx_t$ is the input at the $t$-th
timestep.

The memory cell is computed by
\begin{align}
    \label{eq:lstm_c}
    \vc_t = \vf_t \odot \vc_{t-1} + \vi_t \odot \tanh\left( \mW_c \vx + \mU_c
    \vh_{t-1} + \vb_c \right),
\end{align}
where $\odot$ is an element-wise multiplication. This adaptive leaky integration
of the memory cell allows the LSTM to easily capture long-term dependencies in
the input sequence, and this has recently been widely adopted many works involving 
language models \citep[see, e.g.,][]{sundermeyer2015feedforward}.

The output, or the activation of this LSTM layer, is then computed as
\begin{align*}
    \vh_t = \vo_t \odot \tanh(\vc_t).
\end{align*}

\section{Larger-Context Language Modelling}

In this paper, we aim not at improving the sentence-level probability estimation
$P(S)$ (see Eq.~\eqref{eq:rnn_lm}) but at improving the corpus-level probability
$P(D)$ from Eq.~\eqref{eq:corpus_lm} directly. One thing we noticed at the
beginning of this work is that it is not necessary for us to make the assumption
of mutual independence of sentences in a corpus. Rather, similarly to how we
model a sentence probability, we can loosen this assumption by
\begin{align}
    \label{eq:corpus_lm_better}
    P(D) \approx \prod_{l=1}^L P(S_l|S_{l-n}^{l-1}),
\end{align}
where $S_{l-n}^{l-1} = (S_{l-n}, S_{l-n+1}, \ldots, S_{l-1})$.  $n$ decides on
how many preceding sentences each conditional sentence probability conditions
on, similarly to what happens with a usual $n$-gram language modelling.

From the statistical modelling's perspective, estimating the corpus-level
language probability in Eq.~\eqref{eq:corpus_lm_better} is equivalent to build a
statistical model that approximates
\begin{align}
    \label{eq:rnn_lm_larger}
    P(S_l|S_{l-n}^{l-1}) = \prod_{t=1}^{T_l} p(w_t | w_{<t}, S_{l-n}^{l-1}),
\end{align}
similarly to Eq.~\eqref{eq:rnn_lm}. One major difference from the existing
approaches to statistical language modelling is that now each conditional
probability of a next word is conditioned not only on the preceding words in the
same sentence, but also on the $n-1$ preceding {\em sentences}.

A conventional, count-based $n$-gram language model is not well-suited due to
the issue of data sparsity. In other words, the number of rows in the table
storing $n$-gram statistics will explode as the number of possible sentence
combinations grows exponentially with respect to both the vocabulary size, each
sentence's length and the number of context sentences.

Either neural or recurrent language modelling however does not suffer from this
issue of data sparsity. This makes these models ideal for modelling the {\em
larger-context sentence probability} in Eq.~\eqref{eq:rnn_lm_larger}. More
specifically, we are interested in adapting the recurrent language model for
this.

In doing so, we answer two questions in the following subsections. First, there
is a question of how we should represent the context sentences $S_{l-n}^{l-1}$.
We consider two possibilities in this work. Second, there is a large freedom in
how we build a recurrent activation function to be conditioned on the context
sentences. We also consider two alternatives in this case.

\subsection{Context Representation}
\label{sec:context_rep}

A sequence of preceding sentences can be represented in many different ways.
Here, let us describe two alternatives we test in the experiments.

The first representation is to simply bag all the words in the preceding
sentences into a single vector $\vs \in \left[ 0, 1\right]^{|V|}$. Any element
of $\vs$ corresponding to the word that exists in one of the preceding sentences
will be assigned the frequency of that word, and otherwise $0$. This vector is
multiplied from left by a matrix $\mP$ which is tuned together with all the
other parameters:
\begin{align*}
    \vp = \mP \vs.
\end{align*}
We call this representation $\vp$ a {\em bag-of-words (BoW) context}.

Second, we try to represent the preceding context sentences as a sequence of
bag-of-words. Each bag-of-word $\vs_j$ is the bag-of-word representation of the
$j$-th context sentence, and they are put into a sequence $(\vs_{l-n}, \ldots,
\vs_{l-1})$. Unlike the first BoW context, this allows us to incorporate the
order of the preceding context sentences.  

This sequence of BoW vectors are read by a recurrent neural network which is
separately from the one used for modelling a sentence (see
Eq.~\eqref{eq:rnn_trans}.) We use LSTM units as recurrent activations, and for
each context sentence in the sequence, we get
\begin{align*}
    \vz_t = \phi\left(\vx_t, \vz_{t-1}\right),
\end{align*}
for $t=l-n, \ldots, l-1$.  We set the last hidden state $\vz_{l-1}$ of this
recurrent neural network, to which we refer as a {\em context recurrent neural
network}, as the context vector $\vp$.

\paragraph{Attention-based Context Representation}

The sequence of BoW vectors can be used in a bit different way from the above.
Instead of a unidirectional recurrent neural network, we first use a
bidirectional recurrent neural network to read the sequence. The forward
recurrent neural network reads the sequence as usual in a forward direction, and
the reverse recurrent neural network in the opposite direction. The hidden
states from these two networks are then concatenated for each context sentence
in order to form a sequence of annotation vectors $(\vz_{l-n}, \ldots,
\vz_{l-1})$. 

Unlike the other approaches, in this case, the context vector $\vp$ differs for
each word $w_t$ in the current sentence, and we denote it by $\vp_t$. The
context vector $\vp_t$ for the $t$-th word is computed as the weighted sum of
the annotation vectors:
\begin{align*}
    \vp_t = \sum_{l'=l-n}^{l-1} \alpha_{t,l'} \vz_{l'},
\end{align*}
where the attention weight $\alpha_{t, l'}$ is computed by
\begin{align*}
    \alpha_{t,l'} = \frac{\exp\text{score}\left(\vz_{l'}, \vh_{t}\right)}
    {\sum_{k=l-n}^{l-1} \exp\text{score}\left(\vz_{k}, \vh_{t}\right)}.
\end{align*}
$\vh_t$ is the hidden state of the recurrent language model of the current
sentence from Eq.~\eqref{eq:rnn_out}. The scoring function
$\text{score}(\vz_{l'}, \vh_{t})$ returns a relevance score of the $l'$-th
context sentence with respect to $\vh_t$.

\subsection{Conditional LSTM}
\label{sec:context_feed}

\paragraph{Early Fusion}

Once the context vector $\vp$ is computed from the $n$ preceding sentences, we
need to feed this into the sentence-level recurrent language model. One most
straightforward way is to simply consider it as an input at every time step such
that
\begin{align*}
    \vx = \mE^\top \vw_t + \mW_p \vp,
\end{align*}
where $\mE$ is the word embedding matrix that transforms the one-hot vector
of the $t$-th word into a continuous word vector. This $\vx$ is used by the LSTM
layer as the input, as described in Sec.~\ref{sec:lstm_lm}. We call this
approach an {\em early fusion} of the context into language modelling.

\paragraph{Late Fusion}

In addition to this approach, we propose here a modification to the LSTM such
that it better incorporates the context from the preceding sentences (summarized
by $\vp_t$.) The basic idea is to keep dependencies within the sentence being
modelled ({\em intra-sentence dependencies}) and those between the preceding
sentences and the current sent ({\em inter-sentence dependencies}) separately
from each other. 

We let the memory cell $\vc_t$ of the LSTM in Eq.~\eqref{eq:lstm_c} to model
intra-sentence dependencies. This simply means that there is no change to the
existing formulation of the LSTM, described in
Eqs.~\eqref{eq:lstm_gates}--\eqref{eq:lstm_c}. 

The inter-sentence dependencies are reflected on the interaction between the
memory cell $\vc_t$, which models intra-sentence dependencies, and the context
vector $\vp$, which summarizes the $n$ preceding sentences. We model this by
first computing the amount of influence of the preceding context sentences as
\begin{align*}
    \vr_t = \sigma\left( \mW_r \left(\mW_p \vp\right) + \mW_r \vc + \vb_r\right). 
\end{align*}
This vector $\vr_t$ controls the strength of each of the elements in the context
vector $\vp$. This amount of influence from the $n$ preceding sentences is
decided based on the currently captured intra-sentence dependency structures and
the preceding sentences.

\begin{figure}[t]
    \centering
    \begin{minipage}{0.49\textwidth}
        \centering
        \includegraphics[width=0.95\columnwidth]{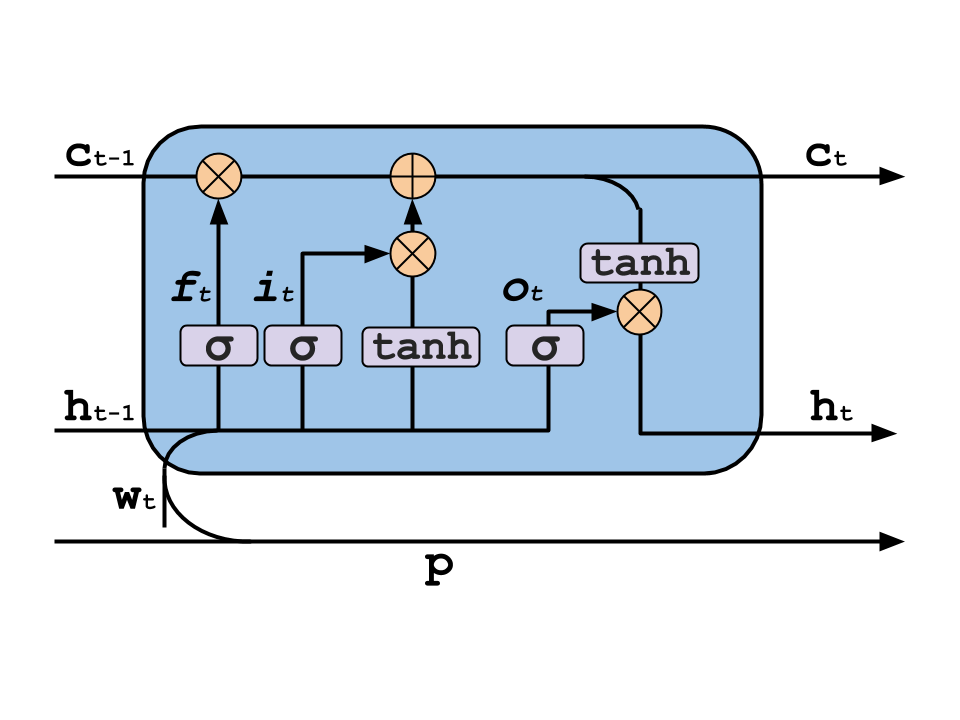}
    \end{minipage}
    \hfill
    \begin{minipage}{0.49\textwidth}
        \centering
        \includegraphics[width=0.95\columnwidth]{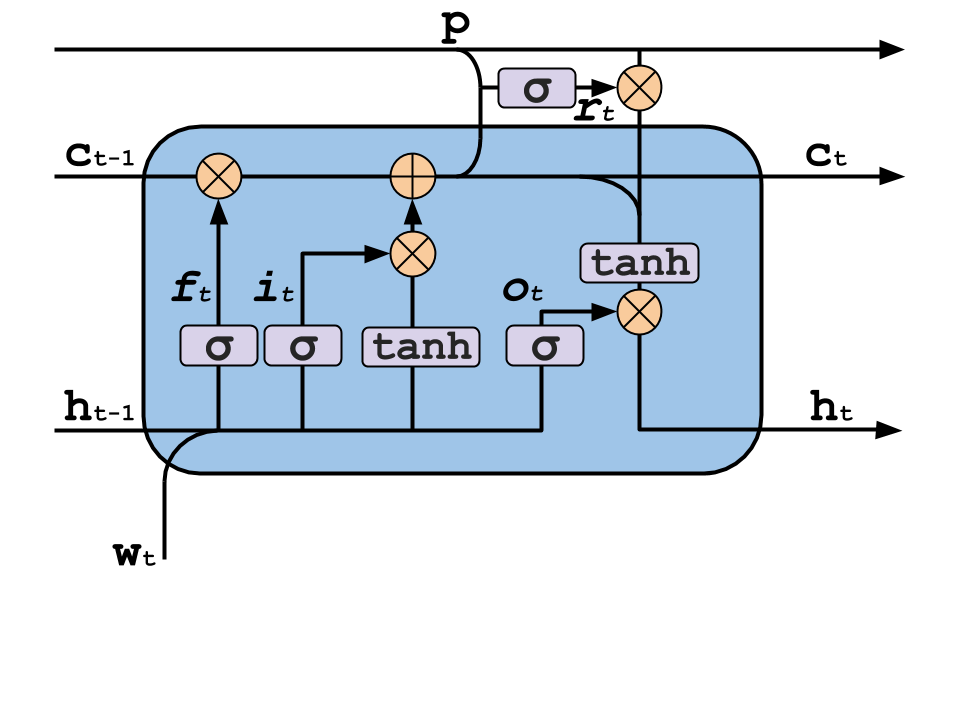}
    \end{minipage}

    \vspace{-6mm}
    \begin{minipage}{0.49\textwidth}
        \centering
        (a)
    \end{minipage}
    \hfill
    \begin{minipage}{0.49\textwidth}
        \centering
        (b)
    \end{minipage}

    \caption{
        Graphical illustration of the proposed (a) early fusion and (b) late fusion.
    }
    \label{fig:fusion}
\end{figure}

This controlled context vector $\vr_t \odot \left(\mW_p \vp\right)$ is then used to compute the
output of the LSTM layer such that
\begin{align*}
    \vh_t = \vo_t \odot \tanh\left(\vc_t + \vr_t \odot \left(\mW_p \vp\right)\right).
\end{align*}
This is illustrated in Fig.~\ref{fig:fusion}~(b).

We call this approach a {\em late fusion}, as the effect of the preceding
context is fused together with the intra-sentence dependency structure in the
later stage of the recurrent activation.

Late fusion is a simple, but effective way to mitigate the issue of vanishing gradient 
in corpus-level language modelling. By letting the context representation flow without 
having to pass through saturating nonlinear activation functions, it provides a linear 
path through which the gradient for the context flows easily. 

\section{Related Work}

\subsection{Context-dependent Recurrent Language Model}

This possibility of extending a neural or recurrent language modeling to
incorporate larger context was explored earlier. Especially,
\citep{mikolov2012context} proposed an approach, called context-dependent
recurrent neural network language model, very similar to the proposed approach
here. The basic idea of their approach is to use a topic distribution,
represented as a vector of probabilities, of previous $n$ {\em words} when
computing the hidden state of the recurrent neural network each time. In doing
so, the words used to compute the topic distribution often went over the
sentence boundary, meaning that this distribution vector was summarizing a part
of a preceding sentence. Nevertheless, their major goal was to use this topic
distribution vector as a way to ``convey contextual information about the
sentence being modeled.'' More recently, \citet{mikolov2014learning} proposed a
similar approach however without relying on external topic modelling.

There are three major differences in the proposed approach from the work by
\citet{mikolov2012context}. First, the goal in this work is to explicitly model
{\em preceding sentences} to better approximate the corpus-level probability
(see Eq.~\eqref{eq:corpus_lm_better}) rather than to get a better context of the
current sentence. Second, \citet{mikolov2012context} use an external method, such as latent
Dirichlet allocation \citep{blei2003latent} or latent semantics analysis
\citep{dumais2004latent} to extract a feature vector, where we learn the whole
model, including the context vector extraction, end-to-end. Third, we propose a
late fusion approach which is well suited for the LSTM units which have recently 
been widely adopted many works involving language models \citep[see,
e.g.,][]{sundermeyer2015feedforward}. This late fusion is later shown to be
superior to the early fusion approach.

Similarly, \citet{sukhbaatar2015end} proposed more recently to use a memory
network for language modelling with a very large context of a hundred to two
hundreds preceding {\em words}. The major difference to the proposed approach is
in the lack of separation between the context sentences and the current sentence
being processed. There are two implications from this approach. First, each
sentence, depending on its and the preceding sentences' lengths, is conditioned
on a different number of context sentences. Second, words in the beginning of
the sentence being modelled tend to have a larger context (in terms of the
number of preceding sentences they are being conditioned on) than those at the
end of the sentence. These issues do not exist in the proposed approach here.

Unlike these earlier works, the proposed approach here uses sentence boundaries
explicitly. This makes it easier to integrate with downstream applications, such
as machine translation and speech recognition, at the decoding level which
almost always works sentence-wise. 

It is however important to notice that these two previous works by
\citet{mikolov2012context} and \citet{sukhbaatar2015end} are not in competition
with the proposed larger-context recurrent language model. Rather, all these
three are orthogonal to each other and can be combined.

\subsection{Dialogue Modelling with Recurrent Neural Networks}

A more similar model to the proposed larger-context recurrent language model is
a hierarchical recurrent encoder decoder (HRED) proposed recently by
\citet{serban2015hierarchical}. The HRED consists of three recurrent neural
networks to model a dialogue between two people from the perspective of one of
them, to which we refer as a speaker. If we consider the last utterance of the
speaker being modelled by the decoder of the HRED, this model can be considered
as a larger-context recurrent language model with early fusion.

Aside the fact that the ultimate goals differ (in their case, dialogue modelling
and in our case, document modelling), there are two technical differences.
First, they only test with the early fusion approach. We show later in the
experiments that the proposed late fusion gives a better language modelling
quality than the early fusion. Second, we use a sequence of bag-of-words to
represent the preceding sentences, while the HRED a sequence of sequences of
words. This allows the HRED to potentially better model the order of the words
in each preceding sentence, but it increases computational complexity (one more
recurrent neural network) and decreases statistical efficient (more parameters
with the same amount of data.)

Again, the larger-context language model proposed here is not competing against
the HRED. Rather, it is a variant, with differences in technical details, that
is being evaluated specifically for document language modelling. We believe many
of the components in these two models are complementary to each other and may
improve each other. For instance, the HRED may utilize the proposed late fusion,
and the larger-context recurrent language model here may represent the context
sentences as a sequence of sequences of words, instead of a BoW context or a
sequence of BoW vectors.

\subsection{Skip-Thought Vectors}

Perhaps the most similar work is the skip-thought vector by
\citet{kiros2015skip}. In their work, a recurrent neural network is trained to
read a current sentence, as a sequence of words, and extract a so-called
skip-thought vector of the sentence. There are two other recurrent neural
networks which respectively model preceding and following sentences.  If we only
consider the prediction of the following sentence, then this model becomes a
larger-context recurrent language model which considers a single preceding
sentence as a context. 

As with the other previous works we have discussed so far, the major difference
is in the ultimate goal of the model. \citet{kiros2015skip} fully focused on
using their model to extract a good, generic sentence vector, while in this
paper we are focused on obtaining a good language model. There are less major
technical differences. First, the skip-thought vector model conditions only on
the immediate preceding sentence, while we extend this to multiple preceding
sentences. The experiments later will show the importance of having a larger
context. Second, similarly to the two other previous works by
\citet{mikolov2012context} and \citet{serban2015hierarchical}, the skip-thought
vector model only implements early fusion.

\subsection{Neural Machine Translation: Conditional Language Modelling}

Neural machine translation is another related approach
\citep{forcada1997recursive,kalchbrenner2013recurrent,cho2014learning,sutskever2014sequence,bahdanau2014neural}.
In neural machine translation, often two recurrent neural networks are used. The
first recurrent neural network, called an encoder, reads a source sentence,
represented as a sequence of words in a source language, to form a context
vector, or a set of context vectors. The other recurrent neural network, called
a decoder, then, models the target translation {\em conditioned on} this source
context.

This is similar to the proposed larger-context recurrent language model, if we
consider the source sentence as a preceding sentence in a corpus. The major
difference is in the ultimate application, machine translation vs. language
modelling, and technically, the differences between neural machine translation
and the proposed larger-context language model are similar to those between the
HRED and the larger-context language model.

Similarly to the other previous works we discussed earlier, it is possible to
incorporate the proposed larger-context language model into the existing neural
machine translation framework, and also to incorporate advanced mechanisms from
the neural machine translation framework. Attention mechanism was introduced by
\citet{bahdanau2014neural} with intention to build a variable-length context
representation in source sentence. In larger-context language model, this
mechanism is applied on context sentences (see Sec.~\ref{sec:context_rep},) and
we present the results in the later section showing that the attention mechanism
indeed improves the quality of language modelling.

\subsection{Context-Dependent Question-Answering Models}

Context-dependent question-answering is a task in which a model is asked to
answer a question based on the facts from a natural language paragraph. The
question and answer are often formulated as filling in a missing word in a query
sentence \citep{hermann2015teaching,hill2015}. This task is closely related to
the larger-context language model we proposed in this paper in the sense that
its goal is to build a model to learn
\begin{align}
    \label{eq:context_qa}
    p(q_k | q_{<k}, q_{>k}, D),
\end{align}
where $q_k$ is the missing $k$-th word in a query $Q$, and $q_{<k}$ and $q_{>k}$
are the context words from the query. $D$ is the paragraph containing facts
about this query. Often, it is explicitly constructed so that the query $q$ does
not appear in the paragraph $D$.

It is easy to see the similarity between Eq.~\eqref{eq:context_qa} and one of
the conditional probabilities in the r.h.s. of Eq.~\eqref{eq:rnn_lm_larger}.  By
replacing the context sentences $S_{l-n}^{l-1}$ in Eq.~\eqref{eq:rnn_lm_larger}
with $D$ in Eq.~\eqref{eq:context_qa} and conditioning $w_t$ on both the
preceding and following words, we get a context-dependent question-answering
model. In other words, the proposed larger-context language model can be used
for context-dependent question-answering, however, with computational overhead.
The overhead comes from the fact that for every possible answer the conditional
probability completed query sentence must be evaluated.

\section{Experimental Settings}

\subsection{Models}

There are six possible combinations of the proposed methods. First, there are
two ways of representing the context sentences; (1) bag-of-words (BoW) and (2) a
sequence of bag-of-words (SeqBoW), from Sec.~\ref{sec:context_rep}.  There are
two separate ways to incorporate the SeqBoW; (1) with attention mechanism (ATT)
and (2) without it.  Then, there are two ways of feeding the context vector into
the main recurrent language model (RLM); (1) early fusion (EF) and (2) late
fusion (LF), from Sec.~\ref{sec:context_feed}. We will denote these six possible
models by
\begin{enumerate}
    \itemsep 0em
    \item RLM-BoW-EF-$n$
    \item RLM-SeqBoW-EF-$n$
    \item RLM-SeqBoW-ATT-EF-$n$
    \item RLM-BoW-LF-$n$
    \item RLM-SeqBoW-LF-$n$
    \item RLM-SeqBoW-ATT-LF-$n$
\end{enumerate}
$n$ denotes the number of preceding sentences to have as a set of context
sentences. We test four different values of $n$; 1, 2, 4 and 8.

As a baseline, we also train a recurrent language model without any context
information. We refer to this model by RLM. Furthermore, we also report the
result with the conventional, count-based $n$-gram language model with the
modified Kneser-Ney smoothing with KenLM~\citep{Heafield-estimate}.

Each recurrent language model uses 1000 LSTM units and is trained with
Adadelta~\citep{zeiler2012adadelta} to maximize the log-likelihood defined as
\begin{align*}
    \LL(\TT) = \frac{1}{K} \sum_{k=1}^K \log p(S_k|S_{k-n}^{k-1}).
\end{align*}
We early-stop training based on the validation log-likelihood and report the
perplexity on the test set using the best model according to the validation
log-likelihood.

We use only those sentences of length up to 50 words when training a recurrent
language model for the computational reason. For KenLM, we used all available
sentences in a training corpus.

\subsection{Datasets}

We evaluate the proposed larger-context language model on three different
corpora. For detailed statistics, see Table~\ref{tab:statistics}.

\begin{table}[t]
\centering
\begin{tabular}{c ||c c|c c|c c}
& \multicolumn{2}{c|}{IMDB} 
& \multicolumn{2}{c|}{BBC} 
& \multicolumn{2}{c}{Penn TreeBank} 
\\
& \# Sentences & \# Words 
& \# Sentences & \# Words 
& \# Sentences & \# Words 
\\
\hline 
\hline 
Training & 930,139 & 21M &  37,207 & 890K & 42,068 & 888K  \\
Validation & 152,987 & 3M & 1,998 & 49K   & 3,370 & 70K \\
Test & 151,987 & 3M &       2,199 &  53K  & 3,761 & 79K \\
\end{tabular}
\caption{Statistics of IMDB, BBC and Penn TreeBank}
\label{tab:statistics}
\end{table}

\paragraph{IMDB Movie Reviews}

A set of movie reviews is an ideal dataset to evaluate many different settings
of the proposed larger-context language models, because each review is highly
likely of a single theme (the movie under review.) A set of words or the style
of writing will be well determined based on the preceding sentences. 

We use the IMDB Move Review
Corpus (IMDB) prepared by \citet{maas2011learning}.\footnote{
    \url{http://ai.stanford.edu/~amaas/data/sentiment/}
} This corpus has 75k training reviews and 25k test reviews. We use the 30k
most frequent words for recurrent language models.


\paragraph{BBC}

Similarly to movie reviews, each new article tends to convey a single theme.  We
use the BBC corpus prepared by \citet{greene06icml}.\footnote{
    \url{http://mlg.ucd.ie/datasets/bbc.html}
} Unlike the IMDB corpus, this corpus contains news articles which are almost
always written in a formal style. By evaluating the proposed approaches on both
the IMDB and BBC corpora, we can tell whether the benefits from larger context
exist in both informal and formal languages. 
We use the 10k most frequent words for recurrent language models.

Both with the IMDB and BBC corpora, we did not do any preprocessing other
than tokenization.\footnote{
    \url{https://github.com/moses-smt/mosesdecoder/blob/master/scripts/tokenizer/tokenizer.perl}
}

\paragraph{Penn Treebank}

We evaluate a normal recurrent language model, count-based $n$-gram language
model as well as the proposed RLM-BoW-EF-$n$ and RLM-BoW-LF-$n$ with varying
$n=1,2,4,8$ on the Penn Treebank Corpus. We preprocess the corpus according to
\citep{mikolov2011extensions} and use a vocabulary of 10k words. 

\section{Results and Analysis}

\begin{figure}[t]
    \centering
    \begin{minipage}{0.95\textwidth}
        \centering
        \includegraphics[width=0.5\columnwidth]{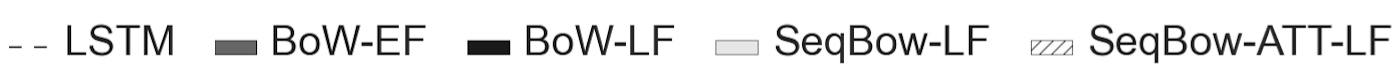}
    \end{minipage}
    \begin{minipage}{0.32\textwidth}
        \centering
        \includegraphics[width=\columnwidth]{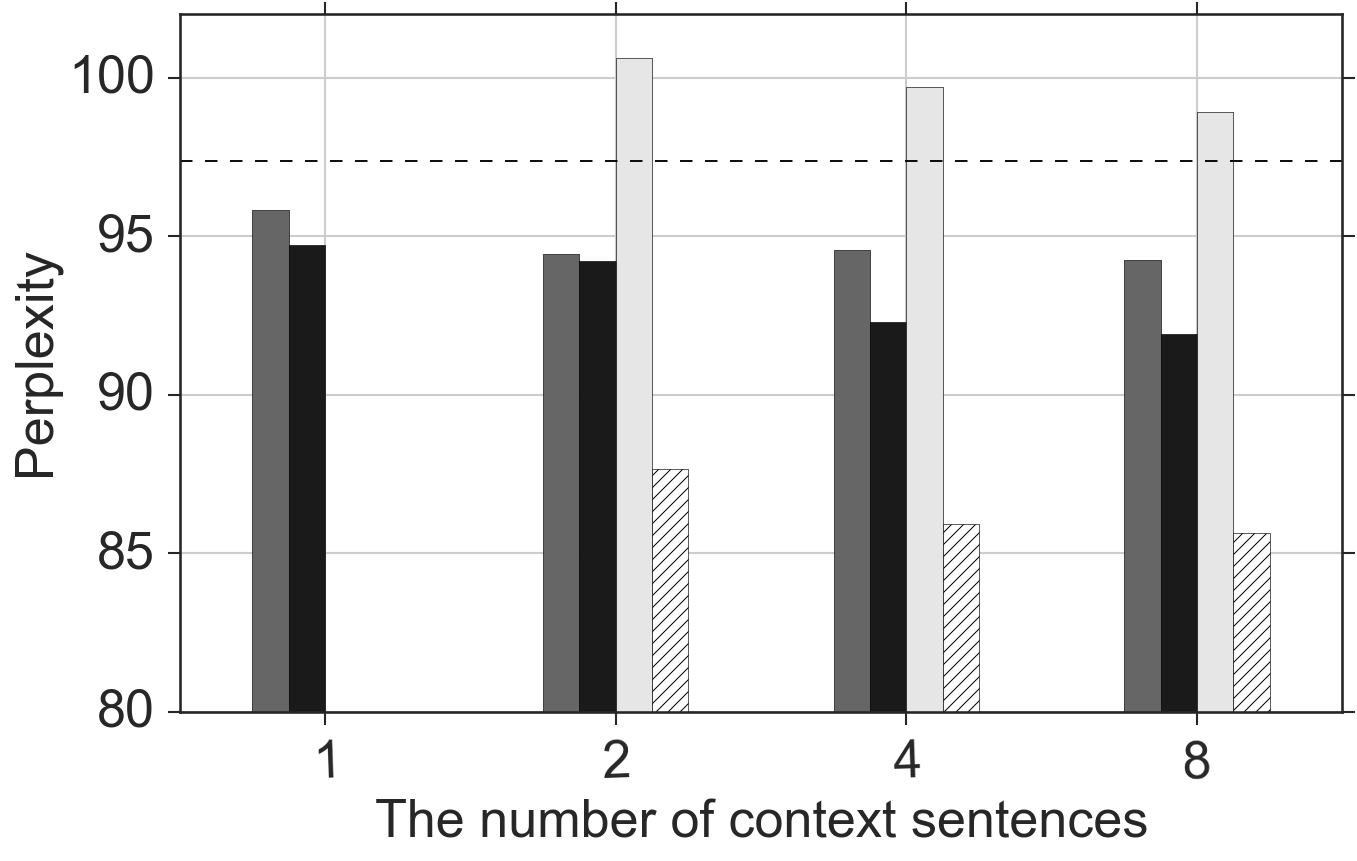}
    \end{minipage}
    \hfill
    \begin{minipage}{0.32\textwidth}
        \centering
        \includegraphics[width=\columnwidth]{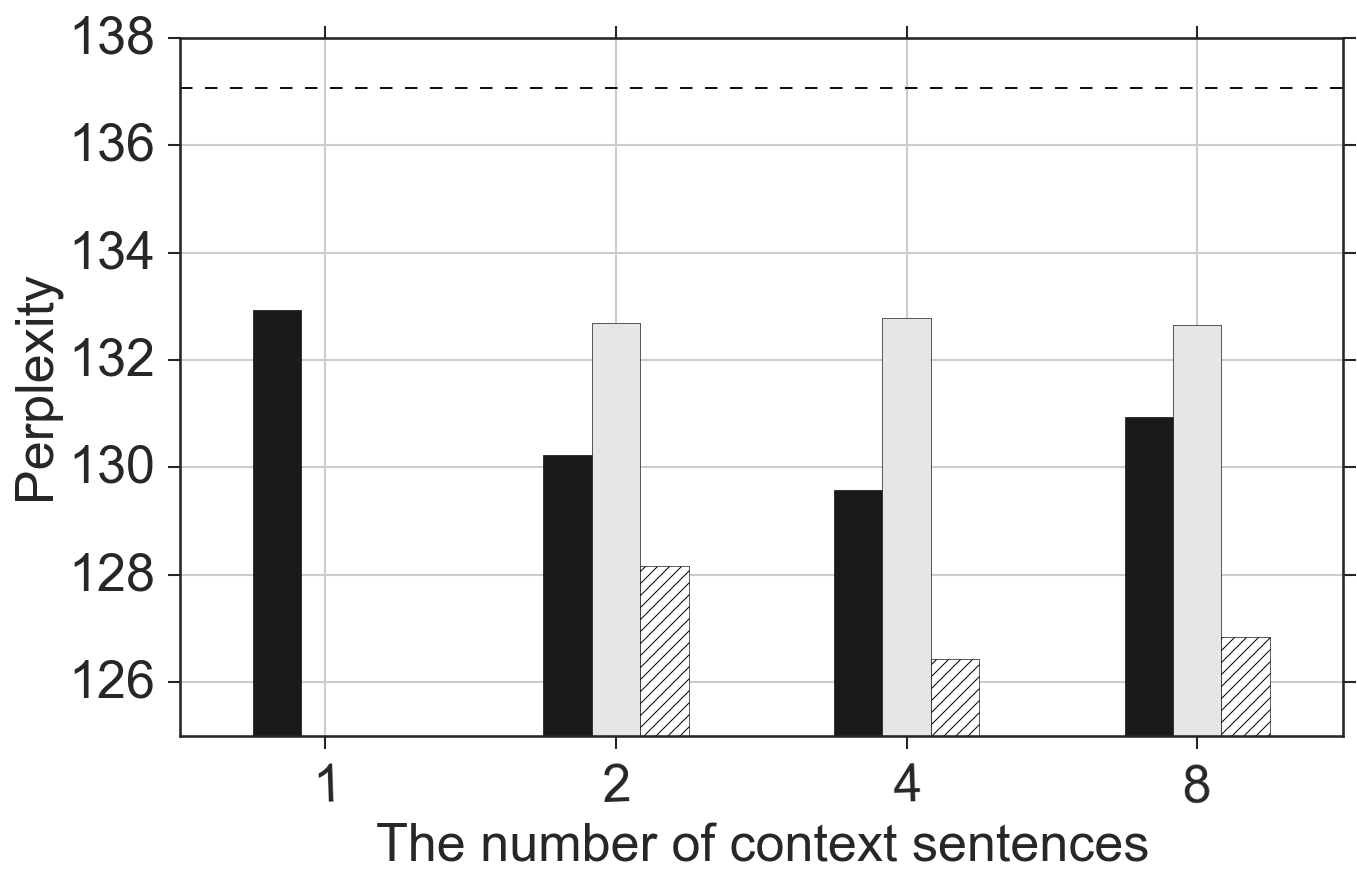}
    \end{minipage}
    \hfill
    \begin{minipage}{0.32\textwidth}
        \centering
        \includegraphics[width=\columnwidth]{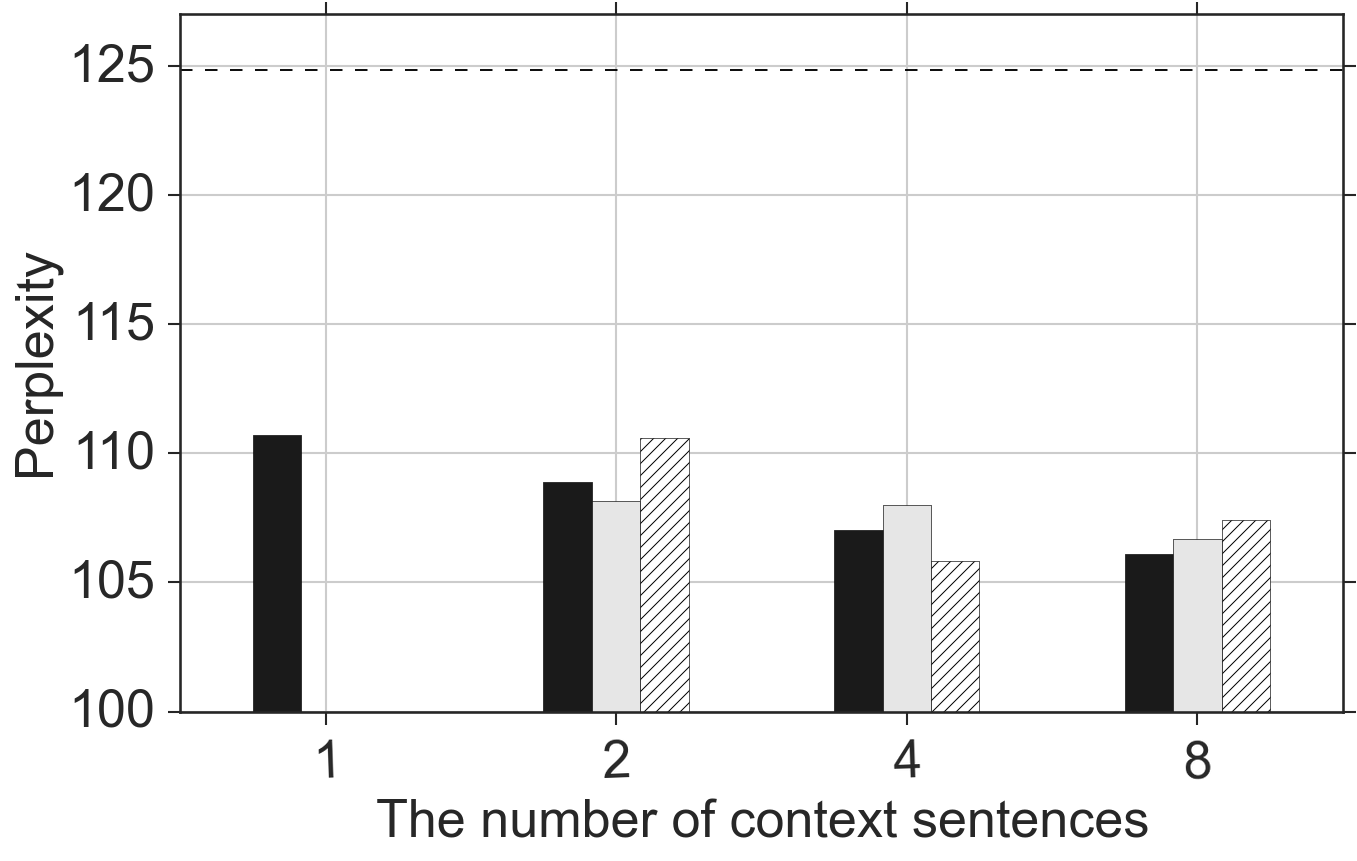}
    \end{minipage}

    \begin{minipage}{0.32\textwidth}
        \centering
        (a) IMDB
    \end{minipage}
    \hfill
    \begin{minipage}{0.32\textwidth}
        \centering
        (b) Penn Treebank
    \end{minipage}
    \hfill
    \begin{minipage}{0.32\textwidth}
        \centering
        (c) BBC
    \end{minipage}

    \caption{
        Corpus-level perplexity on (a) IMDB, (b) Penn Treebank and (c) BBC. The
        count-based 5-gram language models with Kneser-Ney smoothing
        respectively resulted in
        the perplexities of 110.20, 148 and 127.32, and are not shown here. Note
        that we did not show SeqBoW in the cases of $n=1$, as this is equivalent
        to BoW.
    }
    \label{fig:perplexity}
\end{figure}

\subsection{Corpus-level Perplexity}
\label{sec:corpus_result}

We evaluated the models, including all the proposed approaches
(RLM-\{BoW,SeqBoW\}-\{ATT,$\varnothing$\}-\{EF,LF\}-$n$), on the IMDB corpus. In
Fig.~\ref{fig:perplexity}~(a), we see three major trends. First, RLM-BoW, either
with the early fusion or late fusion, outperforms both the count-based $n$-gram
and recurrent language model (LSTM) regardless of the number of context
sentences. Second, the improvement grows as the number $n$ of context sentences
increases, and this is most visible with the novel {\em late fusion}. Lastly, we
see that the RLM-SeqBoW does not work well regardless of the fusion type
(RLM-SeqBow-EF not shown), while after using attention-based model 
RLM-SeqBow-ATT, the performance is greatly improved. 

Because of the second observation from the IMDB corpus, that the late fusion
clearly outperforms the early fusion, we evaluated only
RLM-\{BoW,SeqBoW\}-\{ATT\}-LF-$n$'s on the other two corpora. 

On the other two corpora, PTB and BBC, we observed a similar trend of
RLM-SeqBoW-ATT-LF-$n$ and RLM-BoW-LF-$n$ outperforming the two conventional language 
models, and that this trend strengthened as the number $n$ of the context 
sentences grew. We also observed again that the RLM-SeqBoW-ATT-LF outperforms 
RLM-SeqBoW-LF and RLM-BoW in almost all the cases.

From these experiments, the benefit of allowing larger context to a recurrent
language model is clear, however, with the right choice of the context
representation (see Sec.~\ref{sec:context_rep}) and the right mechanism for
feeding the context information to the recurrent language model (see
Sec.~\ref{sec:context_feed}.) In these experiments, the sequence of bag-of-words 
representation with attention mechanism, together with the late fusion was found 
to be the best choice in all three corpora. 


One possible explanation on the failure of the SeqBoW representation with 
a context recurrent neural network is that it is simply difficult for the context 
recurrent neural network to compress multiple sentences into a single vector. 
This difficulty in training a recurrent neural network to compress a long sequence 
into a single vector has been observed earlier, for instance, in neural machine 
translation~\citep{cho2014properties}. Attention mechanism, which was found to 
avoid this problem in machine translation~\citep{bahdanau2014neural}, is 
found to solve this problem in our task as well. 


\subsection{Analysis: Perplexity per Part-of-Speech Tag}
\label{sec:perp_pos}

Next, we attempted at discovering why the larger-context recurrent language
model outperforms the unconditional recurrent language model. In order to do so,
we computed the {\em perplexity per part-of-speech (POS) tag}. 

We used the Stanford log-linear part-of-speech tagger~\citep[Stanford POS
Tagger,][]{toutanova2003feature} to tag each word of each sentence in the
corpora.\footnote{
    \url{http://nlp.stanford.edu/software/tagger.shtml}
} We then computed the perplexity of each word and averaged them for each tag
type separately. Among the 36 POS tags used by the Stanford POS Tagger, we
looked at the perplexities of the ten most frequent tags (NN, IN, DT, JJ, RB,
NNS, VBZ, VB, PRP, CC), of which we combined NN and NNS into a new tage Noun and
VB and VBZ into a new tag Verb.

\begin{figure}[t]
    \centering
    \begin{minipage}{0.95\textwidth}
        \centering
        \includegraphics[width=\columnwidth]{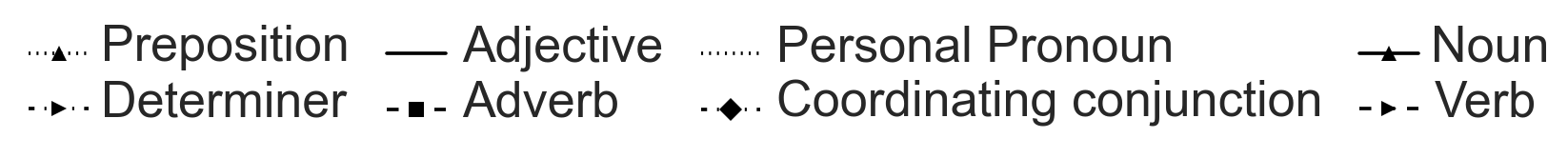}
    \end{minipage}

    \begin{minipage}{0.32\textwidth}
        \centering
        \includegraphics[width=\columnwidth]{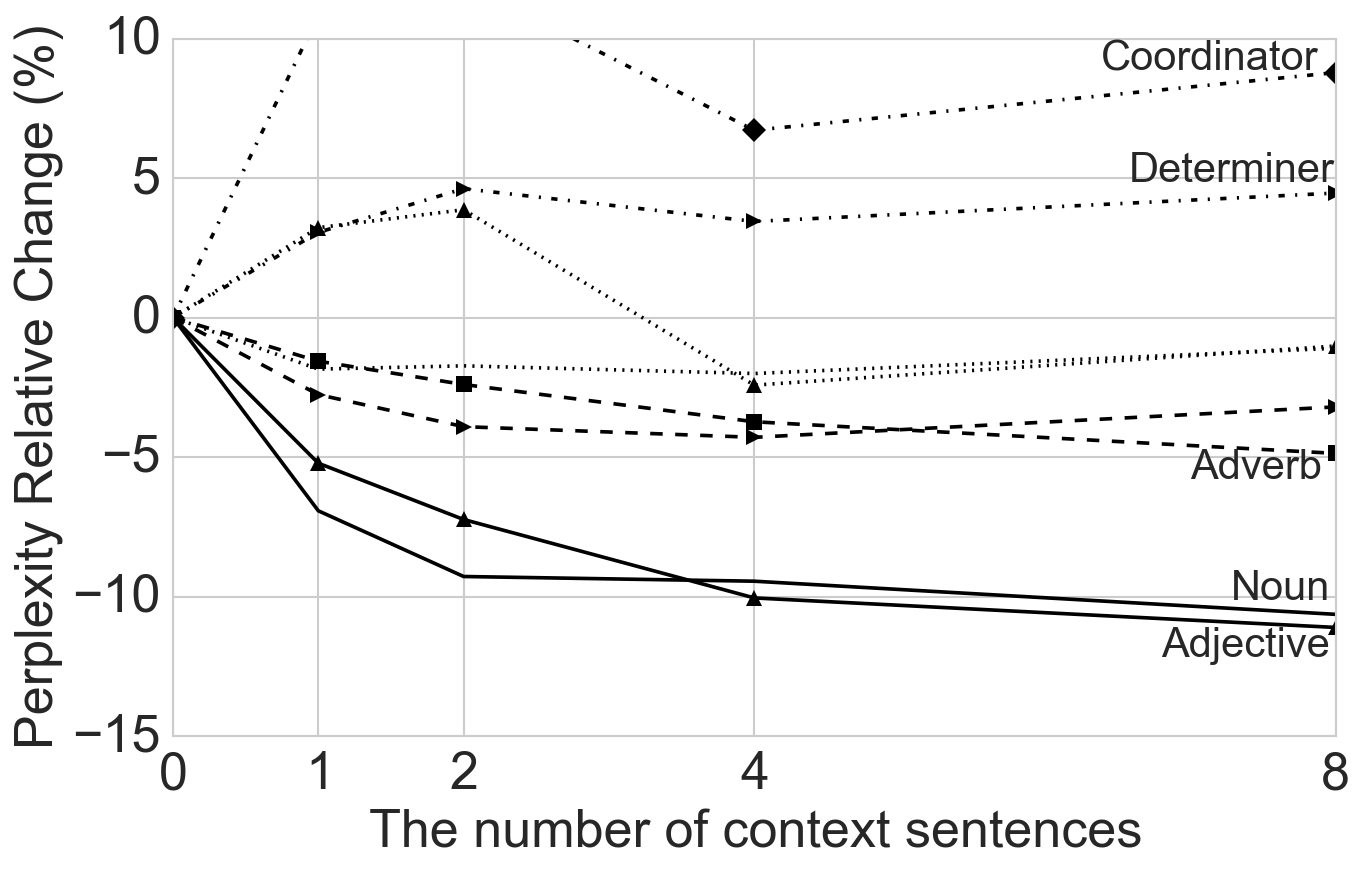}
    \end{minipage}
    \hfill
    \begin{minipage}{0.32\textwidth}
        \centering
        \includegraphics[width=\columnwidth]{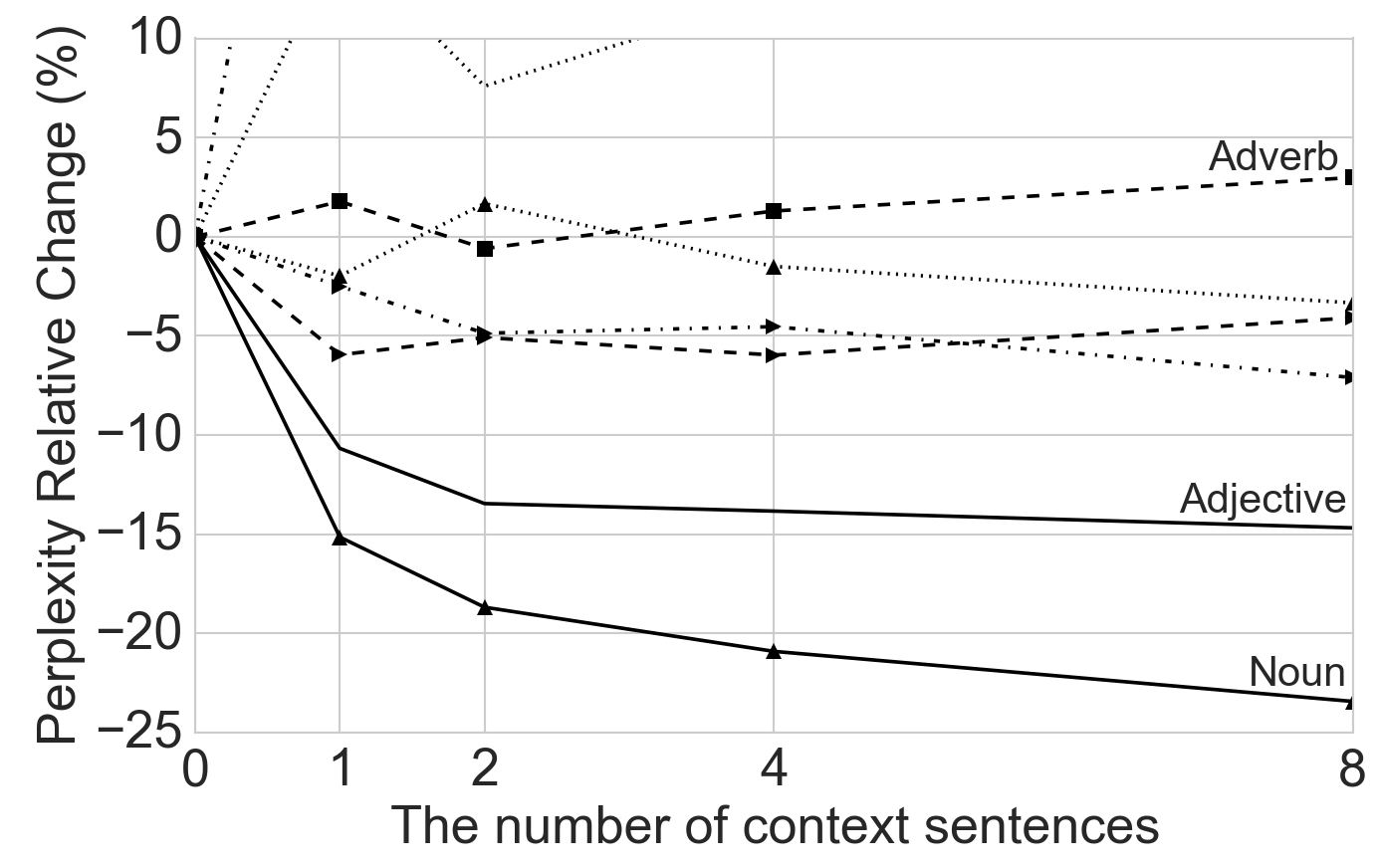}
    \end{minipage}
    \hfill
    \begin{minipage}{0.32\textwidth}
        \centering
        \includegraphics[width=\columnwidth]{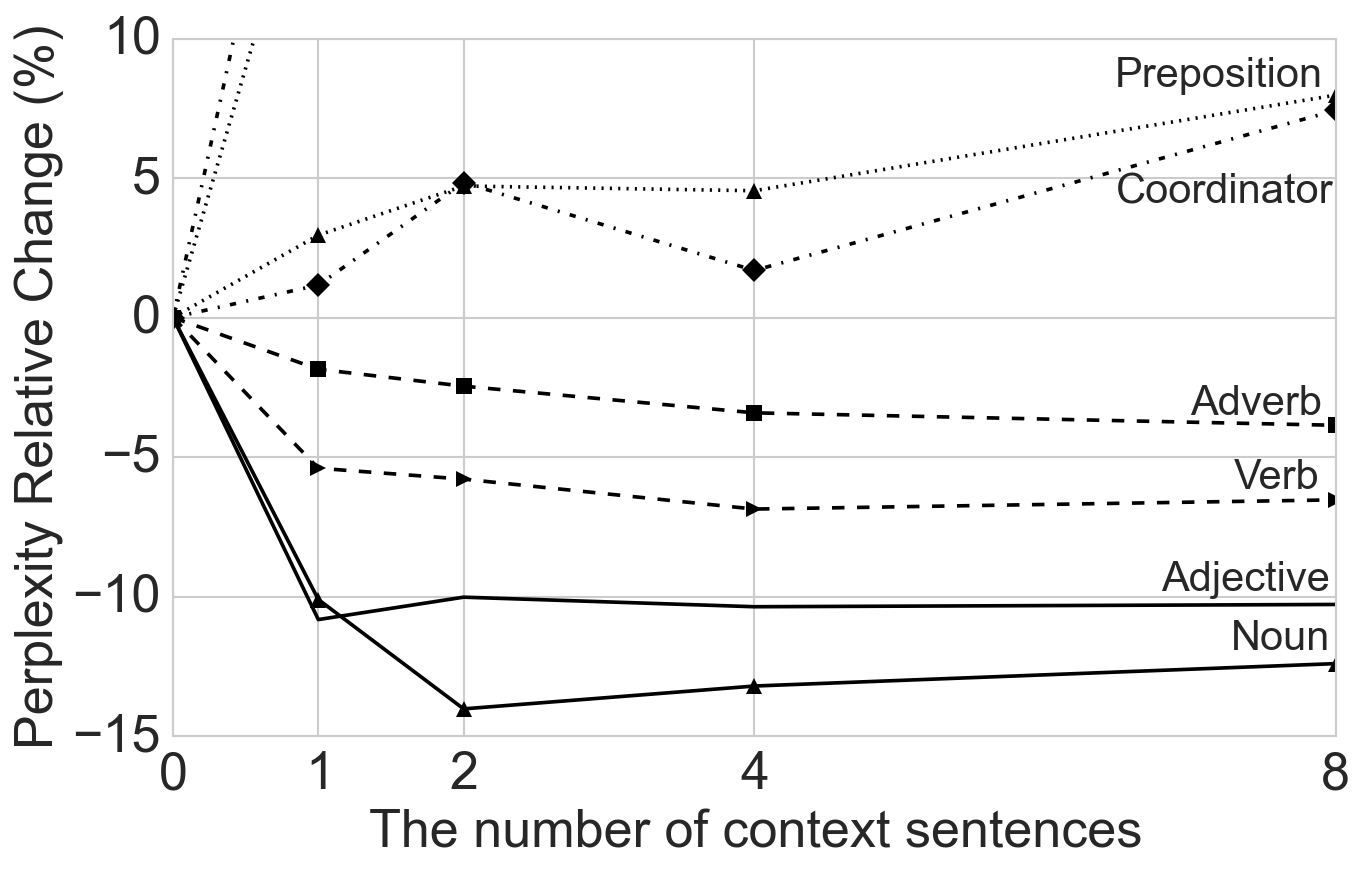}
    \end{minipage}

    \begin{minipage}{0.32\textwidth}
        \centering
        (a) IMDB
    \end{minipage}
    \hfill
    \begin{minipage}{0.32\textwidth}
        \centering
        (b) BBC
    \end{minipage}
    \hfill
    \begin{minipage}{0.32\textwidth}
        \centering
        (b) Penn Treebank
    \end{minipage}
    \begin{minipage}{\textwidth}
        \centering
        (i) RLM-BoW-LF
    \end{minipage}

    \begin{minipage}{0.32\textwidth}
        \centering
        \includegraphics[width=\columnwidth]{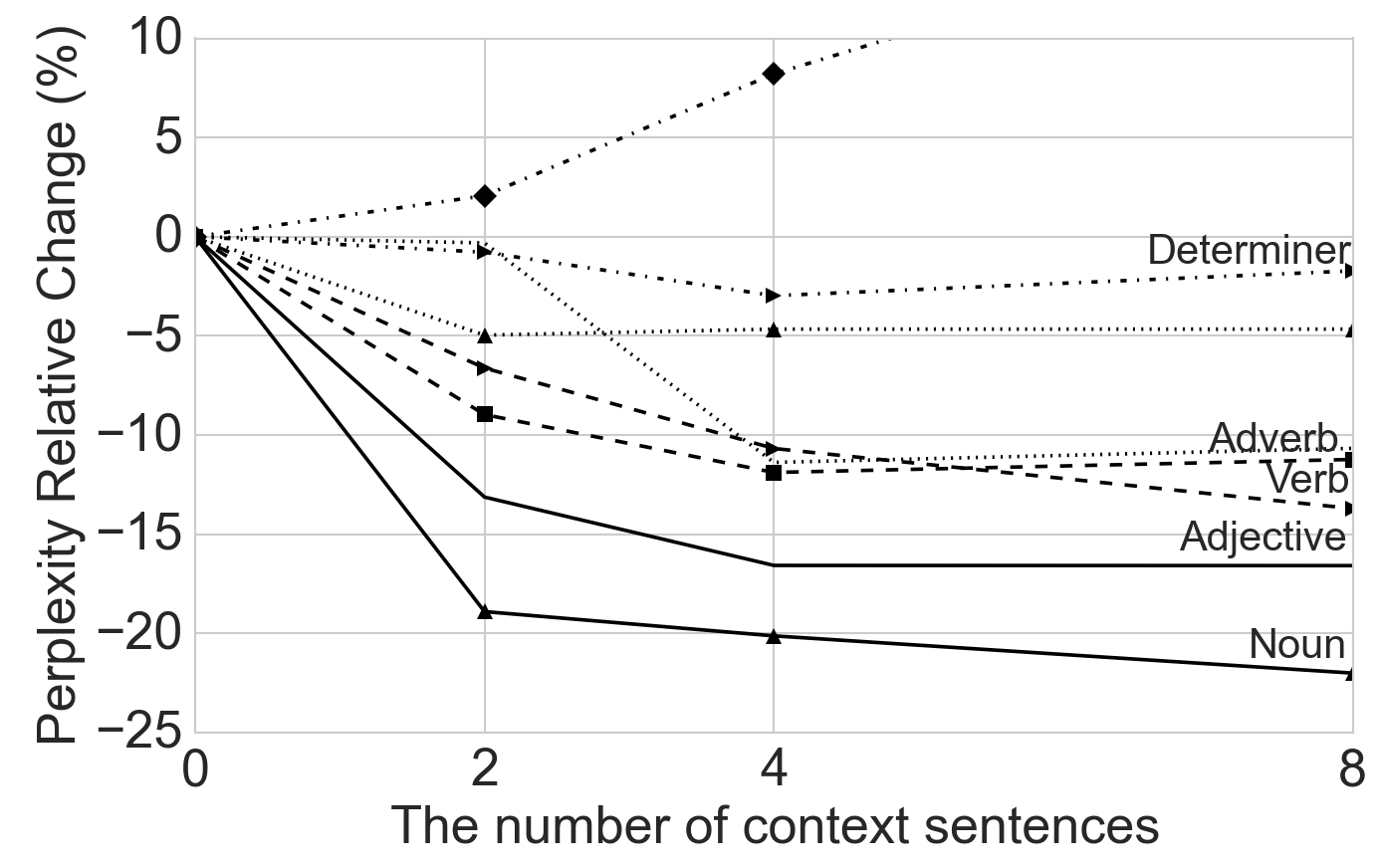}
    \end{minipage}
    \hfill
    \begin{minipage}{0.32\textwidth}
        \centering
        \includegraphics[width=\columnwidth]{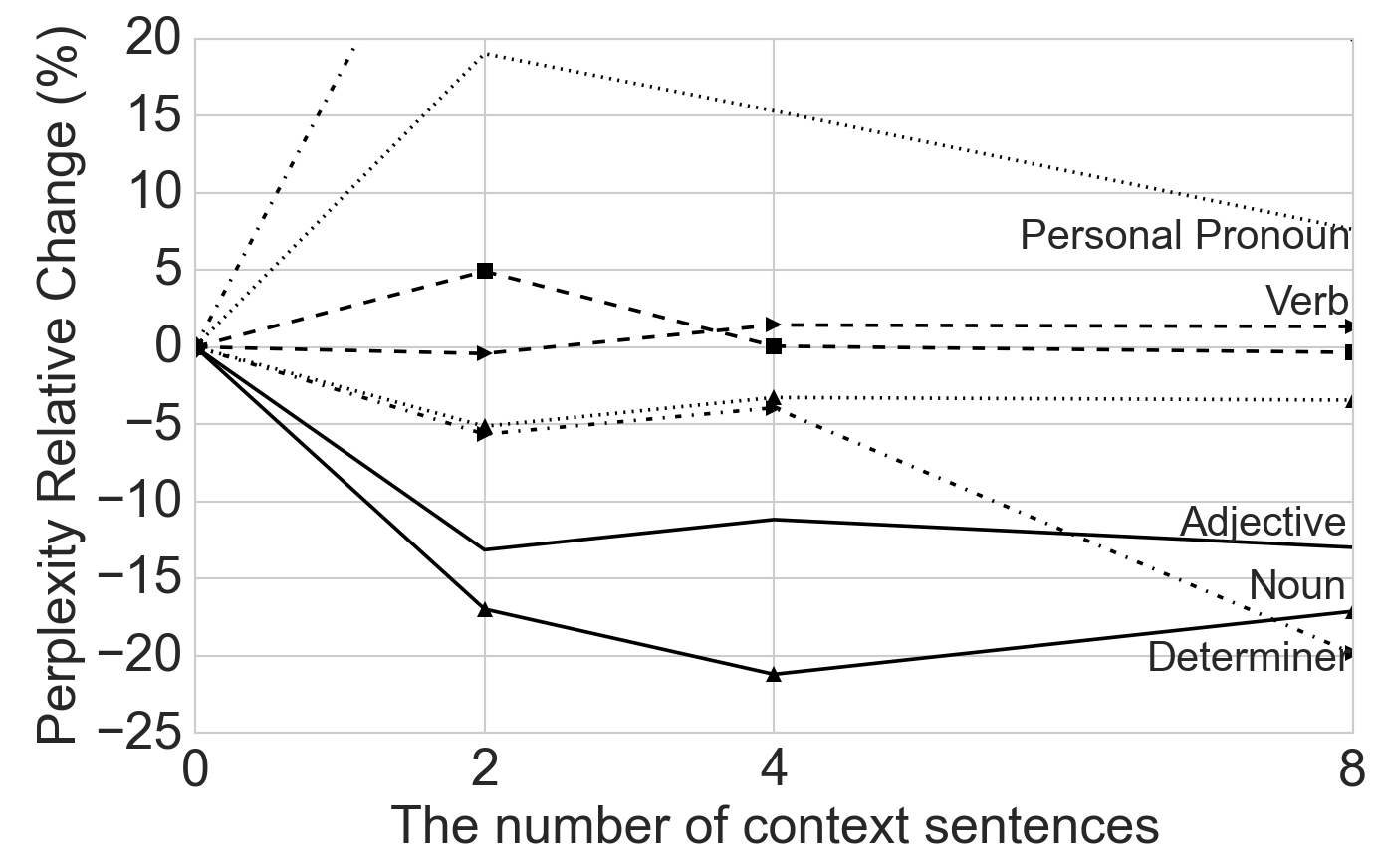}
    \end{minipage}
    \hfill
    \begin{minipage}{0.32\textwidth}
        \centering
        \includegraphics[width=\columnwidth]{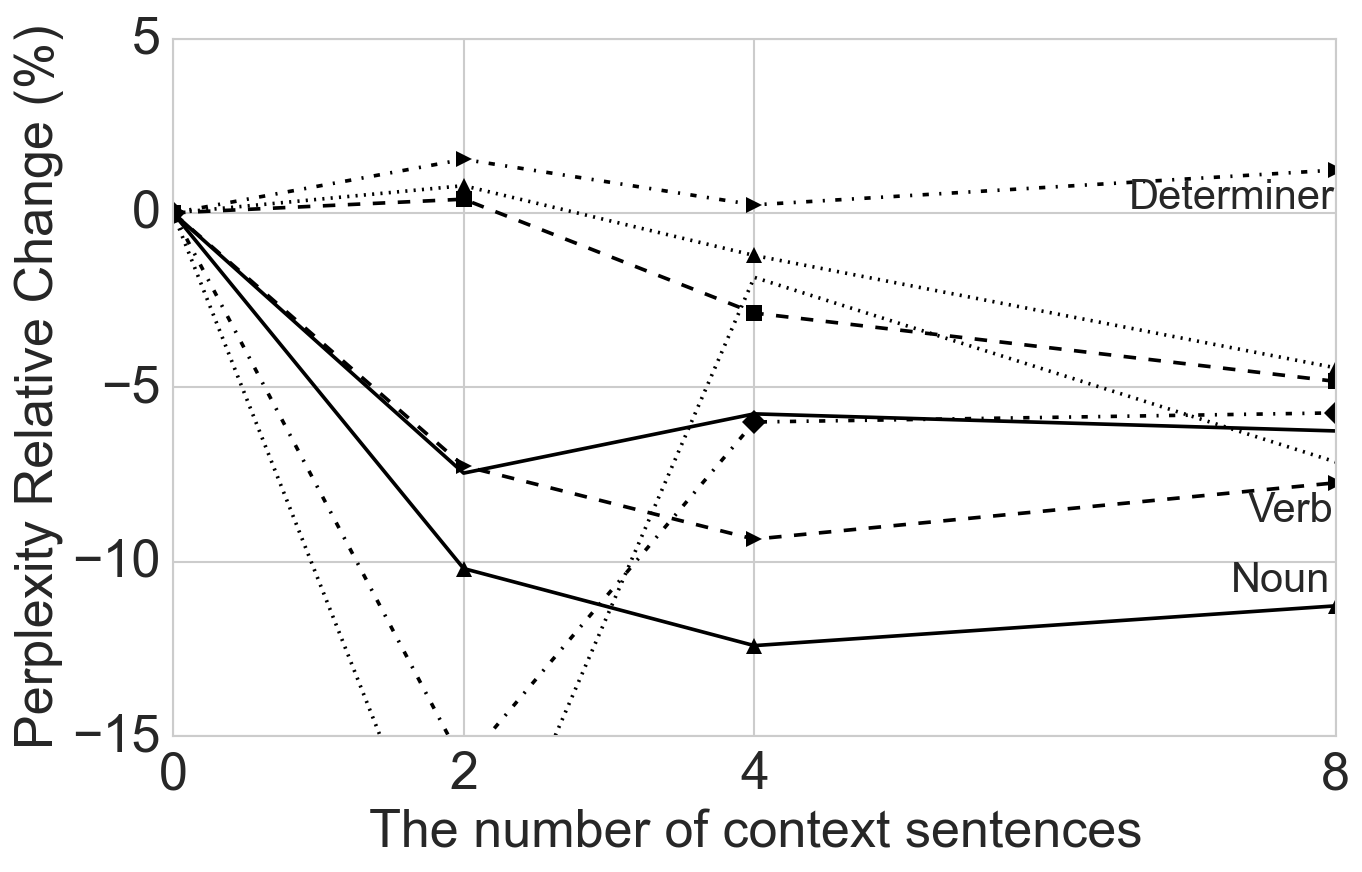}
    \end{minipage}

    \begin{minipage}{0.32\textwidth}
        \centering
        (a) IMDB
    \end{minipage}
    \hfill
    \begin{minipage}{0.32\textwidth}
        \centering
        (b) BBC
    \end{minipage}
    \hfill
    \begin{minipage}{0.32\textwidth}
        \centering
        (b) Penn Treebank
    \end{minipage}
    \begin{minipage}{\textwidth}
        \centering
        (ii) RLM-SeqBoW-ATT-LF
    \end{minipage}

    \caption{
        Perplexity per POS tag on the (a) IMDB, (b) BBC and (c) Penn Treebank 
        corpora.  
    }
    \label{fig:pos}
\end{figure}

We show the results using the RLM-BoW-LF and RLM-SeqBoW-ATT-LF on all three 
corpora--IMDB, BBC and Penn Treebank-- in Fig.~\ref{fig:pos}. We observe that 
the predictability, measured by the perplexity (negatively correlated), grows 
most for nouns (Noun) and adjectives (JJ) as the number of context sentences 
increases. They are followed by verbs (Verb). In other words, nouns, adjectives 
and verbs are the ones which become more predictable by a language model given 
more context.  We however noticed the relative degradation of quality in 
coordinating conjunctions (CC), determiners (DT) and personal pronouns (PRP). 

It is worthwhile to note that nouns, adjectives and verbs are open-class,
content,  words, and conjunctions, determiners and pronouns are closed-class,
function, words \citep[see, e.g.,][]{miller1999knowing}. The functions words
often play grammatical roles, while the content words convey the content of a
sentence or discourse, as the name indicates. From this, we may carefully
conclude that the larger-context language model improves upon the conventional,
unconditional language model by capturing the theme of a document, which is
reflected by the improved perplexity on ``content-heavy'' open-class
words~\citep{chung2007psychological}.  In our experiments, this came however at
the expense of slight degradation in the perplexity of function words, as the
model's capacity stayed same (though, it is not necessary.)

This observation is in line with a recent finding by \citet{hill2015}. They also
observed significant gain in predicting open-class, or content, words when a
question-answering model, including humans, was allowed larger context.


\section{Conclusion}

In this paper, we proposed a method to improve language model on corpus-level by
incorporating larger context. Using this model results in the improvement in
perplexity on the IMDB, BBC and Penn Treebank corpora, validating the advantage
of providing larger context to a recurrent language model. 

From our experiments, we found that the sequence of bag-of-words with attention
is better than bag-of-words for representing the context sentences (see
Sec.~\ref{sec:context_rep}), and the late fusion is better than the early fusion
for feeding the context vector into the main recurrent language model (see
Sec.~\ref{sec:context_feed}). Our part-of-speech analysis revealed that content
words, including nouns, adjectives and verbs, benefit most from an increasing
number of context sentences (see Sec.~\ref{sec:perp_pos}). This analysis
suggests that larger-context language model improves perplexity because it
captures the theme of a document better and more easily. 

To explore the potential of such a model, there are several aspects in which
more research needs to be done. First, the three datasets we used in this paper
are relatively small in the context of language modelling, therefore the
proposed larger-context language model should be evaluated on larger corpora.
Second, more analysis, beyond the one based on part-of-speech tags, should be
conducted in order to better understand the advantage of such larger-context
models. Lastly, it is important to evaluate the impact of the proposed
larger-context models in downstream tasks such as machine translation and speech
recognition.

\subsubsection*{Acknowledgments}

This work is done as a part of the course DS-GA 1010-001 Independent Study in
Data Science at the Center for Data Science, New York University.

\bibliography{widerlm}
\bibliographystyle{iclr2016_conference}

\end{document}